\documentclass[letter, 10pt, conference]{ieeeconf}      
\IEEEoverridecommandlockouts %
\usepackage{times}
\usepackage{graphicx}\graphicspath{{img/}}
\usepackage{epstopdf}
\usepackage{algorithm}
\usepackage{algorithmic}
\usepackage{multirow} 
\usepackage{multicol}
\usepackage{array}
\usepackage{mdwmath}
\usepackage{mdwtab}
\usepackage{rotating}
\usepackage[list=off]{caption}
\usepackage{amssymb}
\usepackage{verbatim}
\usepackage{overpic}
\usepackage{amsmath}
\usepackage{subcaption}
\usepackage[utf8]{inputenc}
\usepackage[T1]{fontenc}
\usepackage[export]{adjustbox}
\usepackage[normalem]{ulem}
\usepackage{siunitx} 
\usepackage{chngpage}

\usepackage{xspace}

\usepackage[acronym]{glossaries}%
\glsdisablehyper 
\makeglossaries%

\newacronym{lfd}{LfD}{Learning from demonstration}
\newacronym{mt}{MT}{machine teaching}
\newacronym{nmse}{NMSE}{normalised mean squared error}
\newacronym{ols} {OLS} {ordinary least squares}
\newacronym{lip}{LiP}{linear-in-the-parameters}

\usepackage{amssymb}
\usepackage{mathtools}
\usepackage{amsmath}
\usepackage{gensymb}

\mathchardef\mhyphen="2D   

\providecommand{\R}     {\mathbb{R}}          
\providecommand{\T}     {\top}                
\providecommand{\I}     {\mathbf{I}}          

\providecommand{\estimated} [1]{\tilde{#1}}

\providecommand{\nd}      {n}                              
\providecommand{\Nd}      {\mathcal{\MakeUppercase{\nd}}}  

\providecommand{\bx}     {\mathbf{x}}         
\providecommand{\nx}     {p}                  
\providecommand{\dimx}   {\mathcal{\MakeUppercase{\nx}}} 

\providecommand{\nu}     {q}                  
\providecommand{\bpi}    {\boldsymbol{\pi}}   

\providecommand{\bphi}  {\boldsymbol{\phi}}               
\providecommand{\bPhi}  {\boldsymbol{\Phi}}               %
\providecommand{\nphi}  {s}                               
\providecommand{\dimphi}{\mathcal{\MakeUppercase{\nphi}}} 
\providecommand{\numD} {\mathcal{N}} 

\providecommand{\wAngle} {\mathcal{\omega}}  

\providecommand{\A}     {\mathcal{A}} 
 
\providecommand{\activationfunction}     {\boldsymbol{\phi}}

\providecommand{\model}{\boldsymbol{\theta}}  

\providecommand{\target}{{\boldsymbol{\theta}^*}}  

\providecommand{\learnt}{\estimated{\boldsymbol{\theta}}}  
\providecommand{\learntaction}{\estimated{\action}}  

\providecommand{\dataset}{\boldsymbol{\mathcal{D}}} 

\providecommand{\argmin}{\mathop{\rm arg~min}\limits}

\providecommand{\searchspaceD}{\boldsymbol{\mathfrak{D}}} 

\providecommand{\action}{u}  

\providecommand{\state}{\boldsymbol{x}}         

\providecommand{\actions}{\boldsymbol{u}}  

\providecommand{\states}{\boldsymbol{X}}         

\providecommand{\noisevar}{\mathcal{\noisestd}^2}  
\providecommand{\noisestd}{\mathcal{\sigma}}  

\providecommand{\reg}      {\mathcal{\lambda}}
\providecommand{\eig}      {b}

\providecommand{\featuref}  {\boldsymbol{\mathcal{\phi}}}    

\providecommand{\triskf}  {\mathcal{\rho}}    

\providecommand{\terrorf}  {\mathcal{\varepsilon}}    

\providecommand{\rmse}  {\mathnormal{E_{RMSE}}} 

\providecommand{\ltwonorm}  {\mathnormal{E_{l2}}}

\providecommand{\length}      {l}   
\providecommand{\mass}      {m}   
\providecommand{\gravity}      {g}   

\renewcommand  {\Nd}      {N}                     

\renewcommand  {\xi}      {q}           
\providecommand{\xii}     {\dot{q}}           
\providecommand{\stiffness}{k}            
\providecommand{\damping}{d}            

\DeclareUnicodeCharacter{0308}{??}

\providecommand{\score}{s}

\FPset{\trialsi}{500} 

\usepackage{xspace}

\providecommand{\eg}{\textit{e.g.,}~} %
\providecommand{\ie}{\textit{i.e.,}~} %
\providecommand*{\sref}[1]{\S\ref{s:#1}}            

\providecommand{\figurename}{Fig.}
\providecommand*{\fref}[1]{\figurename~\ref{f:#1}}  
\providecommand*{\eref}[1]{(\ref{e:#1})}            

\let\labelindent\relax 
\usepackage[inline]{enumitem} 
\setlist{nolistsep}
\providecommand{\il}[1]{\begin{enumerate*}[label=(\roman*)]#1\end{enumerate*}} 
\providecommand{\cl}[1]{\begin{enumerate*}[label=(\alph*)]#1\end{enumerate*}}  

\usepackage[normalem]{ulem}                                        
\usepackage{marginnote}
\usepackage[textwidth=\marginparwidth,colorinlistoftodos]{todonotes}

\colorlet{jb}{red}
\colorlet{mh}{red}
 
\providecommand  {\colorsout}[1]{\bgroup\markoverwith{\textcolor{#1}{\rule[0.5ex]{2pt}{0.4pt}}}\ULon} 

\colorlet{ma}{blue}

\makeatletter\newcommand{\manuallabel}[2]{\def\@currentlabel{#2}\label{#1}}\makeatother

\usepackage[
backend=bibtex,
style=ieee,
url=false,
doi=false,
isbn=false,
doi=true,
eprint=false
]{biblatex}
\usepackage{gensymb}
\usepackage[acronym]{glossaries}
\usepackage{pgfplots}
\usepackage{filecontents}
\usepgfplotslibrary{fillbetween}
\usetikzlibrary{arrows.meta}

\usepackage{tikz}
\usepackage{pgfplots}
\usepgfplotslibrary{statistics}

\title{\LARGE \bf
Training Humans to Train Robots Dynamic Motor Skills}

\author{Marina Y. Aoyama$^{1}$ and Matthew Howard
\thanks{*This work was not supported by any organisation}
\thanks{$^{1}$M. Aoyama and M. Howard are with the Department of Engineering, King's College London, UK. {\tt\small marina.aoyama@kcl.ac.uk}}%
}

\begin{document}%
\maketitle%
\thispagestyle{empty}%
\pagestyle{empty}%

\begin{abstract}
\acrfull{lfd} is commonly considered to be a natural and intuitive way to allow novice users to teach motor skills to robots. However, it is important to acknowledge that the effectiveness of \acrshort{lfd} is heavily dependent on the \emph{quality of teaching}, something that may not be assured with novices. It remains an open question as to the most effective way of guiding demonstrators to produce informative demonstrations beyond \textit{ad hoc} advice for specific teaching tasks. To this end, this paper investigates the use of \emph{machine teaching} to derive an index for determining the quality of demonstrations and evaluates its use in guiding and training novices to become better teachers. Experiments with a simple learner robot suggest that guidance and training of teachers through the proposed approach can lead to up to 66.5\% decrease in error in the learnt skill.
\end{abstract}

\section{INTRODUCTION}
\label{s:intro}

As robots are integrated into many aspects of our life, their use by non-experts is becoming increasingly commonplace \cite{SoniaChernova_AndreaLThomaz2014}. This has led to a growing demand for natural and intuitive means to engender robots with the motor skills to perform useful tasks. {\acrfull{lfd}} is widely considered to be a promising approach to achieve this \cite{BrennaDArgall_SoniaChernova2009,StefanSchaal_AJIjspeert2003}. 

At its core, {\gls{lfd}} can be divided into two stages: \emph{teaching} and \emph{learning}. The teacher, who is usually an expert at performing the task, demonstrates the required motor skills to the robot learner. The latter records data from the demonstration and uses it to compute a skill model using an {\gls{lfd}} algorithm \cite{AudeBillard_SylvainCalinon2008}. While recent years has seen the development of learning algorithms capable of learning a wide range of complex tasks \cite{BrennaDArgall_SoniaChernova2009,Harish_Ravichandar_Athanasios_SPolydoros2020}, their performance is still highly dependent on the \emph{quality of teaching}.

Although human teachers generally try to give informative demonstrations when teaching (rather than, say, randomly sampled data) \cite{MarkHo_MichaelLittman_2016}, recent studies have identified that it is often not intuitive for novice teachers to produce demonstrations of sufficient quality to enable effective learning. For example, in \cite{FaisalKhan_XiaojinZhu2011} none of the human teachers achieved the (provably) optimal strategy for teaching the concept of graspability (\ie whether an object fits in the hand) and \cite{AranSena_MatthewHoward2020} found significant inter-subject variability in ability when teaching a robot a pick and place task. Giving suboptimal demonstrations can result in poor performance of the robot and increase the time, effort and cost of teaching. 

One way to address this, is to seek ways of generating data tailored to optimal performance of the learner robot. For example, \emph{\acrlong{mt}} is an approach that, given knowledge of a learning algorithm, derives training data by solving a bilevel optimisation problem \cite{XiaojinZhu_AdishSingla2018,XiojinZhu2015}. However, so far {\gls{mt}} has only been shown to be tractable for a limited set of problems.


With this in mind, this paper proposes the use of \emph{{\gls{mt}} as a means to guide and train human teachers}, thereby improving the quality of their teaching, while retaining the benefit of their intuition and adaptability in teaching a wide variety of skills. To this end, this research \il{\item defines and solves the {\gls{mt}} problem for teaching a wide class of dynamic motor skills, \item shows how the latter leads to a natural index to measure the quality of the demonstrations and \item evaluates the use of the latter as a means to guide and train human teachers}. Experiments are reported in which the effectiveness of training novice teachers ($n=32$) with the proposed index is evaluated and up to 66.5\% reduction in the error of the robot-learnt skill is seen. 

\section{PROBLEM DEFINITION}%
\label{s:problem}%
%
%
This research aims to optimise the teaching of dynamic motor skills to torque-controlled robots by guiding and training human teachers to produce high quality demonstrations, in the form of data $\dataset$, with bounded teaching effort. Specifically, the skills considered in this paper are those representable as a closed loop controller
\begin{equation}\label{e:SkillModel} 
    \action = \bpi(\state,\model)
\end{equation}
where $\bpi(\state)$ maps the system state $\state \in \R^\dimx$ to the desired action $\action\in\R$, and $\model\in\R^\dimphi$ are the skill parameters.

The demonstration set $\dataset$ consists of a list of states $[\state_1,...,\state_\numD]=\states \in \R^{\dimx\times\numD}$ and the corresponding list of actions $(\action_1,...,\action_\numD)^\T=\actions \in \R^\numD$ provided by the teacher to demonstrate the skill. These may take the form of trajectories or key frames sampled from the target behaviour. 
	
Using this data, the learner then forms a model of the skill 
\begin{equation}
	\learntaction = \bpi(\state,\learnt) \label{e:estimatedpi(x)}
\end{equation}
with which to reproduce the behaviour by estimating the parameters through some learning algorithm
\begin{equation}
\learnt = \A (\dataset) \label{e:A(D)}.
\end{equation}

A promising approach to deriving high quality teaching data, beyond simply relying on the intuition of demonstrators, is to make use of {\gls{mt}} techniques. The latter use knowledge of the learning algorithm \eref{A(D)} and the target skill parameters $\target$ to derive conditions on training data such that learning performance is optimised. In general, {\gls{mt}} can be expressed as a bi-level optimisation problem and several formulations are possible. In this paper, the formulation
\begin{align}
	\dataset&=\arg\min_{\dataset \in \searchspaceD} 
\triskf (\learnt, \target) \label{e:MTBilevel1} \\ 
	&\mathrm{s.t.}\quad \learnt = \A (\dataset) \label{e:MTBilevel2} \\
	&\mathrm{and}\quad \terrorf (\dataset)\le n_B. \label{e:MTBilevel3}
\end{align}
is used \cite{GambellaClaudio_GhaddarBissan_2020}. Here, the \emph{teacher's} problem \eref{MTBilevel1} and \eref{MTBilevel3} is to find the training set $\dataset$ from the space of possible data sets $\searchspaceD$ that minimises a \emph{teaching risk function} $\triskf (\learnt, \target)$ subject to a constraint on the effort expended represented by the \emph{teaching effort function} $\terrorf (\dataset)$. The latter reflects the fact that in real-world robot {\gls{lfd}}, typically providing demonstrations is costly (\eg in terms of the teacher's time, equipment wear and tear and computational burden). In this paper, a fixed effort budget $n_B$ is assumed, consisting of the maximum number of data points in $\dataset$.
The \emph{learner's} problem \eref{MTBilevel2} is to optimise the model of the skill (\ie the machine learning problem) based on the data\footnote{Note that, an implicit assumption here is that the model can be taught \emph{only through demonstrations} and not by directly passing $\target$ to the learner. This reflects the fact that $\target$ is typically only \emph{implicitly} known to the demonstrator, through their own skill in performing the task.} $\dataset$. 

When the target model $\target$ is explicitly known and \eref{MTBilevel1}-\eref{MTBilevel3} can be solved, the machine teacher can create the optimal training data $\dataset$. However, in many cases, \il{\item the target model is only known \emph{implicitly} and \item the {\gls{mt}} problem is intractable,} especially when the motor skills are complex or the learning algorithm does not have a closed-form solution. Therefore, this study examines if and how \emph{the optimality benefits of machine teaching can be combined with the versatility of human teaching}. Specifically, the idea is to exploit human teachers' (often implicit) knowledge of a broad range of motor skills, but shape their teaching strategies toward those deemed optimal according to \gls{mt}, by training the human teachers. It is anticipated that this can lead to a significant improvement in the overall quality of teaching, for example, enabling humans to transfer optimal teaching of one motor skill to another, including those for which the {\gls{mt}} problem cannot be directly solved.

To this end, the aims of this paper are twofold, namely, \il{\item to quantify the potential improvements to learner performance that may be achieved by optimising demonstrations through {\gls{mt}}, and \item investigating the extent to which {\gls{mt}} may be used to inform the training of non-experts to become skilled in teaching motor skills to robots}. 

\section{METHOD}%
\label{s:method}%
In this section, the approach taken to formulating the {\gls{mt}} problem for the teaching of dynamic motor skills is described alongside the key modelling assumptions.

\subsection{Modelling the Learner} \label{s:learner}
The learner’s goal is to learn a model of the target skill  $\target$ from demonstrations $\dataset$ using a learning algorithm $\A$. As a simple but general model, in this paper, it is assumed that the learner approximates the target motor skill \eref{SkillModel} with a {\gls{lip}} model
\begin{equation}\label{e:LiPSkillModel} 
\learntaction = \learnt^\T\featuref(\state)
\end{equation}
where $\featuref(\state)\in\R^\dimphi$ is a vector of features or basis functions.

To learn \eref{LiPSkillModel} without overfitting, it is assumed that the learner forms the approximation through ridge regression, that is, by minimising the loss function 
\begin{equation}\label{e:LipLoss} 
    \min\limits_{\learnt \in \R^\dimphi}\sum_{i=1}^{\numD}\frac{1}{2}(\learnt^{\T}\bphi_{i} -\action_i)^2+\frac{\lambda }{2}\left \| \learnt  \right \|^2
\end{equation}
where $\lambda$ is the regularisation parameter and $\left \| \model  \right \|$ is the Mahalanobis norm of $\model$. The closed-form solution of \eref{LipLoss} is
\begin{equation} \label{e:LipSolve} 
    \learnt=(\bPhi  \bPhi^\T + \lambda \I)^{-1}\bPhi \actions
\end{equation}
where $\I\in\R^{\dimphi\times\dimphi}$ is the identity matrix and $\bPhi \in \R^{\dimphi\times\numD}$ is a feature matrix mapped from $\states$ through the basis functions $\featuref(\state)$. 

\subsection{Modelling the Teacher}
\label{s:teacher}

The teacher's primary goal is to produce the demonstration set $\dataset$ that enables the learner to estimate a model \eref{LiPSkillModel} that reproduces the the target skill \eref{SkillModel} as closely as possible. To this end, the teacher's model of the skill is
\begin{equation}\label{e:action_noise} 
\action = \target^\T\featuref(\state).
\end{equation}
The teacher's goal is to minimise the teaching risk function
\begin{equation}
\label{e:RiskFunction}
\triskf (\learnt, \target) = \mathbb{E}[(\learnt-\target)^{\T}(\learnt-\target)]
\end{equation}
where $\mathbb{E}$ denotes the expectation over the error in $\learnt$ model when there is Gaussian noise (assumed to originate from sensors or demonstrator errors) $\epsilon\sim\numD(0,\noisestd)$ on $\actions$.

In line with the formulation of {\gls{mt}} outlined in \sref{problem}, the teacher must teach the target skill with a fixed budget of teaching effort. In the following, the teaching budget \eref{MTBilevel3} is set as
\begin{equation} \label{e:TeachingBudget}
    \terrorf (\dataset) = \numD_{TD}
\end{equation}
where $\numD_{TD}$ is the \emph{teaching dimension} for the problem, defined as the minimum number of training items required to teach the target model to the learner. For the learner defined in \sref{learner}, the teaching dimension is $\dimphi$ \ie the dimensionality of the target model \cite{JiLiu_XiaojinZhu2016}.

\subsection{Machine Teaching Problem}
\label{s:mt_problem}
Combining \eref{LipSolve}, \eref{RiskFunction} and \eref{TeachingBudget}, the {\gls{mt}} problem used in this paper is therefore
\begin{align}\label{e:Optimisation}
\dataset&= \argmin_{\dataset \in \searchspaceD} \mathbb{E}[(\learnt-\target)^{\T}(\learnt-\target)] \\
\mathrm{s.t.\quad}  \learnt&=(\bPhi  \bPhi^\T + \lambda \I)^{-1}\bPhi \actions\quad
\label{e:mt3}
\mathrm{and}\quad \terrorf (\dataset) = \dimphi.
\end{align}
The solution to \eref{Optimisation}-\eref{mt3} is the optimal data set with which to teach the learner robot, and is therefore the gold standard of teaching to which human teachers should aspire. However, it should be noted that in general this solution is \il{\item non-unique, and \item hard to find in closed form for cases where $\dimphi>2$}. With this in mind, rather than prescribing a specific way of teaching (\ie $\dataset$) when training teachers, the strategy taken in this paper is to find the \emph{conditions for demonstration optimality} under \eref{Optimisation}-\eref{mt3}. These conditions can then be used to guide teachers toward improving their teaching ability.

\subsection{Condition for Optimal Demonstrations}
\label{s:2dmodel}
In this section, the conditions for demonstration optimality are derived. As noted in \sref{mt_problem}, the general solution to  for $\dimphi>2$ is challenging to find, so the approach taken here is to first perform a simple analysis of the conditions for optimal teaching under \eref{Optimisation}-\eref{mt3} to gain insight into the problem. It will then be shown that this approach is consistent with the analytically solvable case of $\dimphi=2$.

\subsubsection{Optimal Teaching for $\dimphi>2$}
\label{s:dimphigt2}
A consequence of the formulation \eref{Optimisation}-\eref{mt3} is that the feature matrix  for the optimal data set is, by definition, square \ie $\bPhi\in\R^{\dimphi\times\dimphi}$. If it is further assumed that \il{\item the feature vectors are normalised (\ie $0\le\|\bphi(\state)\|\le1\ \forall\ \state$) and \item for the optimal data set regularisation is not needed (\ie $\lambda=0$)}, the learning problem \eref{LipSolve} can be reduced to
\begin{equation}\label{e:ReducedLipSolve}
\learnt=\bPhi^{-1}\actions.
\end{equation}
A condition for solving \eref{ReducedLipSolve} is that the inverse exists, or equivalently
\begin{equation}\label{e:ReducedLipSolveCondition}
	\det\bPhi\ne0.
\end{equation}
This leads to the following condition for optimal teaching:
\begin{center}\fbox{\noindent\textbf{Condition:} \emph{Optimal demonstrations maximise $\lvert\det\bPhi\rvert$.}}\end{center}
Even when $\lambda\ne0$, choices of $\bPhi$ with determinant close to zero risk numerical instability.

\subsubsection{{ Optimal Teaching for $\dimphi=2$}}
In the case that $\dimphi=2$, a general expression of the feature matrix $\bPhi\in\R^{2\times2}$, assuming that the feature vectors are normalised (\ie $0\le\|\bphi(\state)\|\le1\ \forall\ \state$), is

\begin{equation} \label{e:Phi2d} 
    \bPhi = \left(\begin{array}{cc}\bphi_1 & \bphi_2\end{array}\right) = 
          \left(\begin{array}{cc} 1 & \cos\wAngle \\ 0 & \sin\wAngle\end{array}\right)
\end{equation}
where $\wAngle$ is the angle between the feature vectors $\bphi_1$ and $\bphi_2$. 

Using (\ref{e:LipSolve}), \eref{RiskFunction} can be written \cite{YazidMAIHassan_Performanceof2010}
\begin{equation}
\label{e:trisk_mse}
\triskf (\learnt, \target) = \noisevar \sum_{i=1}^{2} \frac{\eig_i}{(\eig_i+\reg)^2}+\reg^2\target^{\T}(\bPhi\bPhi^{\T}+\reg \I)^{-2}\target
\end{equation}
where $\eig_i$ is the $i$th eigenvalue of $\bPhi^{\T}\bPhi$ and $\noisevar$ is the variance of the noise on the training data (see \eref{action_noise}). 
Since $\lambda$ is small the second term, it is neglected here to give
\begin{equation}
\label{e:trisk_mse1}
\triskf (\learnt, \target) = \noisevar \sum_{i=1}^{2} \frac{\eig_i}{(\eig_i+\reg)^2}
\end{equation}
where
\begin{equation} \label{e:eig2d} 
	b_1=1+\sqrt{1-\sin^2\wAngle}\quad\mathrm{and}\quad b_2=1-\sqrt{1-\sin^2\wAngle}.
\end{equation}
From
\eref{trisk_mse1}-\eref{eig2d},
\begin{equation}\label{e:dpdw21} 
\frac{d\triskf}{d\wAngle} = \dfrac{4 \noisevar \cos\wAngle\sin\wAngle\left(\left(2\lambda+1\right)\sin^2\wAngle-2\lambda^3-3\lambda^2-2\lambda\right)}{(\sqrt{1-\sin^2\wAngle}-\lambda-1)^3(\sqrt{1-\sin^2\wAngle}+\lambda+1)^3}.
\end{equation}
It can be shown that $\wAngle^* = \pi/2$ satisfies $d\triskf/d\wAngle = 0$ and $d^2\triskf/d\wAngle^2>0$.

Moreover, it can be shown that for the feature matrix \eref{Phi2d}
\begin{equation} \label{e:detPhi}
\lvert\det\bPhi\rvert = \lvert\sin(\wAngle)\rvert
\end{equation}
which is maximised for $\wAngle=\wAngle^*$, hence lending support to the condition defined in \sref{dimphigt2}.

\subsection{Teaching Quality Index}
\label{s:index}
Having defined the conditions for optimal demonstration data in the preceding section, it is now possible to design a strategy for improving human teaching.
Specifically, the proposed \emph{teaching quality index} is
\begin{equation}\label{e:score}
    s=100\lvert\det\bPhi\rvert
\end{equation}
which is a score out of 100 (assuming normalised feature vectors). 

The latter can be used either as an \emph{evaluation} (\ie to \emph{assess} teaching ability) or as \emph{guidance} (\ie \emph{feedback} to teachers) to enable them to improve their teaching.

\section{EVALUATION} 
\label{s:evaluation}
In this section, the teaching framework in \sref{method} is applied to teaching of motor skills to a torque-controlled pendulum as a simple, dynamically-controlled learner robot. First, the learner's performance when given data meeting the optimality condition derived in \sref{2dmodel} is compared to that of sub-optimal demonstrations in simulation. Then, the efficacy of using the teaching quality index proposed in \sref{index} as a tool to guide and train human teachers is evaluated and compared to natural (\ie non-guided) teaching. 

\subsection{Role of Demonstration Optimality}
\label{s:simulation}
This evaluation aims to verify that use of demonstrations that meet the optimality condition proposed in \sref{method} maximise learner performance in learning dynamic motor skills. 

The system state is described by $\bx =(\xi,\xii)^\T$ where $\xi$ and $\xii$ represent the angular position and velocity, respectively. The pendulum is actuated by the torque $\action=\tau$ applied by the motor at the pivot. For simplicity, the pendulum length $\length$ is $1\,m$, its mass $\mass$ is $1\,kg$ and gravity $\gravity$ is taken as $9.81\,m/s^2$. No torque limit is set in this simulation and the sampling rate is $10\,kHz$ throughout.

In this evaluation, the goal of teaching is to get the robot to learn controllers representing simple `motor skills', namely,
\begin{enumerate*}[label=S\arabic*] 
	\item\label{i:s1} \emph{undamped oscillation}, and,
	\item\label{i:s2} \emph{rapid movement without overshoot} (akin to critical damping).
\end{enumerate*}
Note that, both of these skills can be represented by a {\gls{lip}} controller of the form \eref{SkillModel}, where $\model =(\stiffness,\damping)^\T$ and $\activationfunction(\bx)=(\sin\xi,\xii)^\T$. For instance, undamped oscillation is achieved with $\model=(\gravity/\length,0)^\T$. To teach these skills, demonstration data $\dataset=\{\state_\nd,\action_\nd\}_{\nd=1}^\Nd$ is provided to the robot, containing sample states $\state_\nd$ and actions $\action_\nd$. The robot learns by applying \eref{LipSolve} to this data with $\lambda=10^{-6}$.

\pgfplotsset{compat=1.10}

\setlength{\textfloatsep}{5pt}
\begin{figure}[t!]
\centering
\begin{tikzpicture}[trim left=-0.5cm]
    \begin{axis}[ylabel shift = -8pt,xlabel shift = -5pt,ymode=log,thick,smooth,no markers,ymin=0.01, ymax=10^7,width=0.27\textwidth,
            height=0.2\textwidth, legend style={nodes={scale=0.5, transform shape}},xlabel={$\wAngle(rad)$},
    ylabel={$\rmse$},label style={font=\footnotesize},tick label style={font=\footnotesize},xmax=1.58,xmin=0
]

        \addplot+[black] table[x=x,y=y] {s1_015.dat};
        \addplot+[name path=A,draw=none] table[x=x,y=high] {s1_015.dat};
        \addplot+[name path=B,draw=none] table[x=x,y=low] {s1_015.dat};
        \addplot[black!40] fill between[of=A and B];
        
        \addplot+[red] table[x=x,y=y] {s1_010.dat};
        \addplot+[name path=A2,draw=none] table[x=x,y=high] {s1_010.dat};
        \addplot+[name path=B2,draw=none] table[x=x,y=low] {s1_010.dat};
        \addplot[red!40] fill between[of=A2 and B2];

        \addplot+[solid,gray] table[x=x,y=y] {s2_010.dat};
        \addplot+[name path=A3,draw=none] table[x=x,y=high] {s2_010.dat};
        \addplot+[name path=B3,draw=none] table[x=x,y=low] {s2_010.dat};
        \addplot[gray!40] fill between[of=A3 and B3];

    \end{axis}
    \begin{axis}[axis x line*=top, axis y line=none, ylabel shift = -8pt,xlabel shift = -5pt,ymode=log,thick,smooth,no markers,ymin=0.01, ymax=10^7,width=0.27\textwidth,
            height=0.2\textwidth, ,xlabel={\ref{i:RMSEvsW}},xticklabels={},xtick={0.5,1.0, 1.5},
    label style={font=\footnotesize},xmax=1.58,xmin=0
]\end{axis}
    
\end{tikzpicture}
\begin{tikzpicture}[trim left=-0.7cm]
    \begin{axis}[xlabel shift = -5pt,thick,smooth,no markers,width=0.27\textwidth,
            height=0.2\textwidth, legend style={nodes={scale=0.5, transform shape},draw=none},xlabel={$\wAngle (rad)$},
    ylabel={$\ltwonorm$},label style={font=\footnotesize},tick label style={font=\footnotesize},xmax=1.58,xmin=0,ymin=0
]

        \addplot+[black] table[x=x,y=y] {s1_015_l2.dat};
        \addplot+[name path=A,draw=none,forget plot] table[x=x,y=high] {s1_015_l2.dat};
        \addplot+[name path=B,draw=none,forget plot] table[x=x,y=low] {s1_015_l2.dat};
        \addplot[black!40,forget plot] fill between[of=A and B];
        
        \addplot+[red] table[x=x,y=y] {s1_010_l2.dat};
        \addplot+[name path=A2_l2,draw=none,forget plot] table[x=x,y=high] {s1_010_l2.dat};
        \addplot+[name path=B2_l2,draw=none,forget plot] table[x=x,y=low] {s1_010_l2.dat};
        \addplot[red!40,forget plot] fill between[of=A2_l2 and B2_l2];
 
        \addplot+[solid,gray] table[x=x,y=y] {s2_010_l2.dat};
        \addplot+[name path=A3,draw=none,forget plot] table[x=x,y=high] {s2_010_l2.dat};
        \addplot+[name path=B3,draw=none,forget plot] table[x=x,y=low] {s2_010_l2.dat};
        \addplot[gray!40,forget plot] fill between[of=A3 and B3];

        \legend{{\ref{i:s1}, $\noisestd=0.15$},%
        	    {\ref{i:s1}, $\noisestd=0.1~$},%
        	    {\ref{i:s2}, $\noisestd=0.1~$}}
    \end{axis}
    \begin{axis}[axis x line*=top, axis y line=none, ylabel shift = -8pt,xlabel shift = -5pt,ymode=log,thick,smooth,no markers,ymin=0.01, ymax=10^7,width=0.27\textwidth,
            height=0.2\textwidth, ,xlabel={\ref{i:El2vsW}},xticklabels={},xtick={0.5,1.0, 1.5},
    label style={font=\footnotesize},xmax=1.58,xmin=0
]\end{axis}
\end{tikzpicture}
\caption{Error in \cl{\item\label{i:RMSEvsW} behaviour reproduced by the learner and \item\label{i:El2vsW} learnt model} when teaching \ref{i:s1} ($\noisestd\in{0.1, 0.15}$) and \ref{i:s2} ($\noisestd=0.1)$. Shown are mean$\pm0.1$s.d. over \trialsi\ trials of learning.}%
\label{f:rmse_l2}%
\end{figure}

To simulate the demonstration process, training data is created by sampling pairs of data points $\dataset=\{(\state_1,\action_1),(\state_2,\action_2)\}$ where $\state_1=\activationfunction^{-1}((1,0)^\T)$ and $\state_2$ is computed through \eref{Phi2d}. The corresponding action demonstrations are computed by applying \eref{action_noise}. This data is used to teach the learner robot through the process described in \sref{2dmodel}. To evaluate the effect of teaching with data of varying quality, a series of independent data sets are generated with varying \il{\item angles between the feature vectors($\pi/36\le\wAngle\le\pi/2\, rad$, \ie $0.087\le\lvert\det\bPhi\rvert\le1.0$) and \item noise levels, \ie $\noisestd\in\{0.05,0.1,0.15\}$}.
Teaching is repeated for \trialsi\ times per choice of $\wAngle$ and $\noisestd$. To evaluate the quality of learning the learner is then made to produce a sample trajectory exemplifying the skill, and (i) the root mean squared error against that produced by the target controller
\vspace{-3mm}
\begin{equation} \label{e:rmse}
\setlength{\abovedisplayskip}{5pt}
\setlength{\belowdisplayskip}{5pt}
	\rmse = \frac{1}{S}\sum_{s = 0}^{S}\sqrt{(\bx_s-\hat{\bx}_s)^\T(\bx_s-\hat{\bx_s})}
\end{equation}
and (ii) the difference between the target and learnt model 
\begin{equation} \label{e:l2norm}
\setlength{\abovedisplayskip}{5pt}
\setlength{\belowdisplayskip}{5pt}
\ltwonorm = \sqrt{(\model-\target)^\T(\model-\target)}
\end{equation}
are computed. In the results reported below, the sample trajectory starts from initial state $\bx =(\pi/2,0)^\T$, and runs for $3\,s$ (\ie $S=30000$ steps).
\fref{rmse_l2}\ref{i:RMSEvsW} and \ref{i:El2vsW} are plots of the $\rmse$ and the $\ltwonorm$ against the angle between the feature vectors $\wAngle$, respectively. 
As can be seen, for all noise levels, as $\wAngle$ approaches $\pi/2\,rad$ (\ie determinant of $\bPhi$ is maximised) the error in the reproduced trajectory rapidly drops. 
\fref{traj} shows representative trajectories reproduced using the learnt model $\learnt$ when learning from data generated with $\wAngle=\pi/18$, $\wAngle=5\pi/18$ and $\wAngle=\pi/2\,rad$.
As can be seen, for the latter case, the reproduced trajectory almost exactly overlaps the desired trajectory. In contrast, the behaviour learnt with suboptimal choices of $\wAngle$, while reproducing some aspects of the behaviour, has lower accuracy. These results suggest that the selection of high-quality training data (\ie meeting the optimality condition derived in \sref{method}) is crucial to effective learning.
\subsection{Training Human Teachers}
\label{s:experiment}
The second experiment\footnote{This study is conducted under the approval of the King's College London Research Ethics Committee, Ref.: MRSP-20/21-21429. Informed consent was obtained from all experimental  participants. The data collected for this research is open access, with accreditation, from {\tt\small http://doi.org/[link to be created upon acceptance]}.} aims to evaluate how giving guidance using the proposed teaching quality index can help human teachers provide better demonstrations and improve learner performance. The hypotheses are that \il{\item the quality of demonstrations produced by human teachers improves when they receive guidance from machine teaching, \item human teachers \emph{retain} their improved skill in teaching after guidance is removed and \item human teachers can \emph{generalise} from guidance provided for the teaching of one skill to improve their teaching of another, different skill}. 

\begin{figure}[t]
    \centering
    \includegraphics[width=0.4\textwidth]{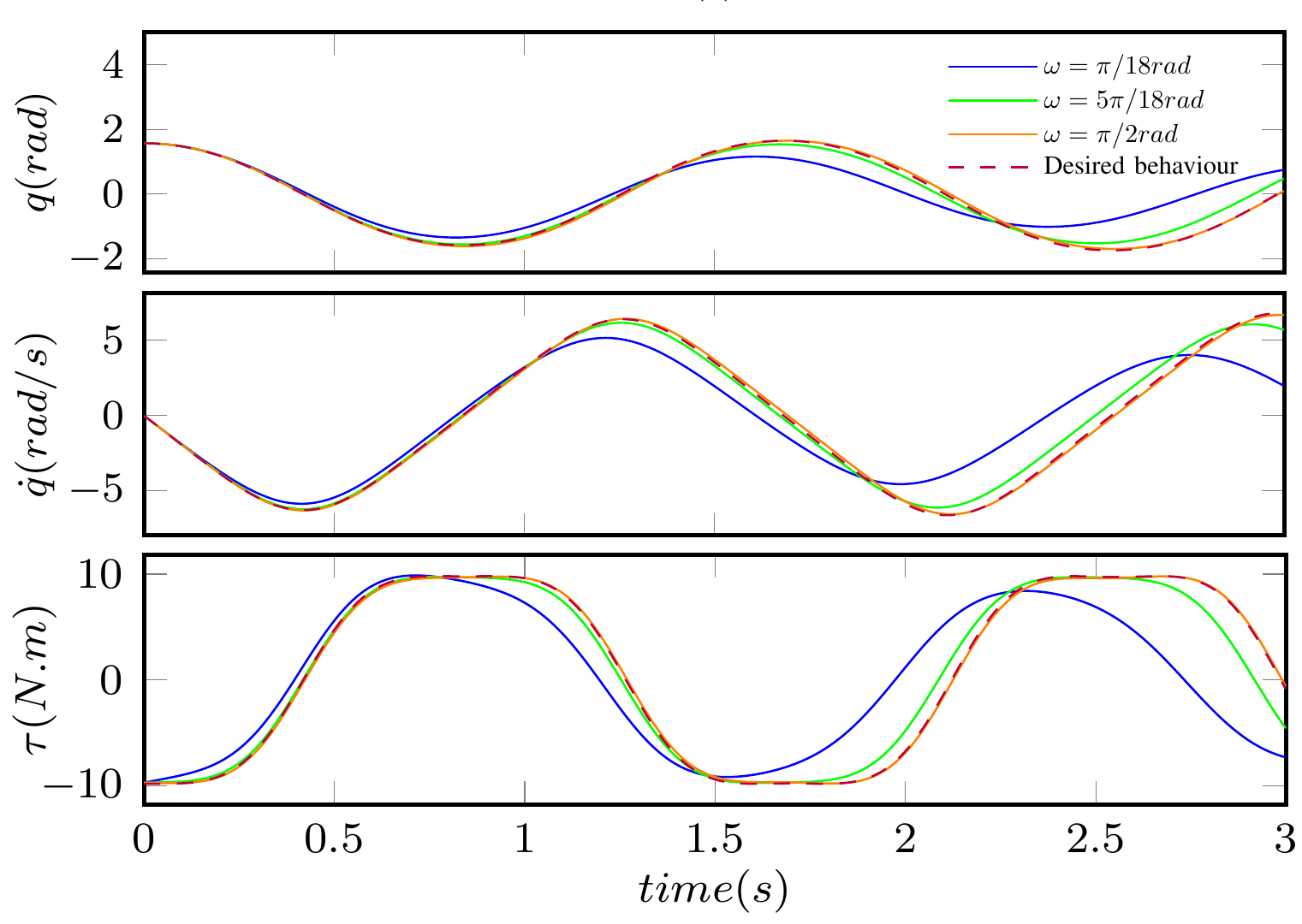}
    \caption{Sample trajectories reproduced using the learnt 
      model $\learnt$ when learning from data generated with $\wAngle=\pi/18$, $\wAngle=5\pi/18$, $\wAngle=\pi/2\,rad$ and noise $\noisestd=0.1$.}
      \label{f:traj}
\end{figure}

For this evaluation, the robot learner is the same torque-controlled pendulum system as described in \sref{simulation}. Participants are asked to teach the two motor skills previously described (\ie \ref{i:s1} and \ref{i:s2}). Demonstrations are provided through the online, interactive experimental interface shown in \fref{exp_interface}. 

The experiment reported here is designed as an interventional study, in which the teaching behaviour of a target group, that are given guidance via machine-teaching, is compared to that of a control group. To this end, the experiment consists of two types of trials: non-guided and guided trials. In both kinds of trial, participants are shown a visualisation of the target behaviour and asked to provide demonstrations in form of two via points (\ie states) that they consider informative for learning the behaviour. In the non-guided trials, participants are asked to select these without any feedback or guidance. In the guided trials, they select the demonstrations and are then given a score $0\le\score\le100$ using \eref{score}. Note that, participants can re-select via points and get feedback as many times as they wish when choosing their demonstrations, but can provide only one pair of points to the learner. Once selected, the corresponding actions \eref{action_noise} with $\epsilon\sim\numD(0,0.1)$ are provided to the learner alongside the chosen points. 

\begin{figure}[t]
    \centering
    \includegraphics[width=0.3\textwidth]{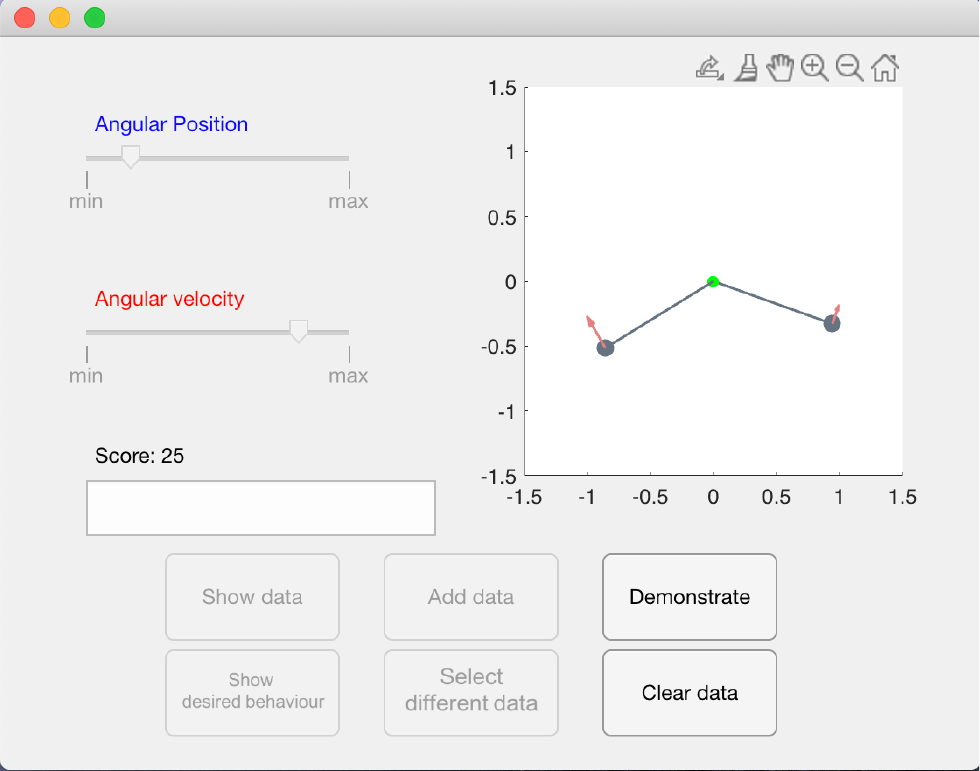}
    \caption{Experimental interface. Participants provide demonstrations as via points (\ie angular position and velocity) using the sliders (top left). The visualisation of selected points is displayed in the white box in the upper right corner. }
    \label{f:exp_interface}
\end{figure}

The experimental procedure is as follows. Study participants($n=32$) are recruited and randomly assigned to either the target or control group. Participants are shown video instructions\footnote{The instruction video is submitted as supplementary material.} on how to teach the robot learner using the interface, and then proceed to the main experiment which consists of six phases:

\begin{enumerate}[label=P\arabic*]
	\item\label{i:phase1}\textit{Skill 1, no guidance.} Participants give demonstrations to teach \ref{i:s1} without guidance.
   \item\label{i:phase2}\textit{Skill 2, no guidance.} Participants give demonstrations to teach \ref{i:s2} without guidance.
   \item\label{i:phase3}\textit{Skill 1, guidance.} Participants in the target group give demonstrations to teach \ref{i:s1} with guidance (\ie given the score $\score$). Participants in the control group repeat \ref{i:phase1}.
   \item\label{i:phase4}Participants repeat \ref{i:phase2}.
   \item\label{i:phase5}Participants repeat \ref{i:phase1}.
   \item\label{i:phase6}Participants repeat \ref{i:phase2}.
\end{enumerate}

A \textit{post-hoc} test on data from \ref{i:phase1} and \ref{i:phase2} indicated no significant difference in teaching behaviour (\ie determinant of $\bPhi$ and $\rmse$) between the control and target groups ($p>0.05$).

To evaluate improvement in teaching ability, and the resultant quality of learning, \il{\item the determinant of $\bPhi$, and \item the $\rmse$ of a trajectory produced from the learner is computed from each participant's demonstrations at every phase of the experiment}. The changes in the determinant of $\bPhi$ and the $\rmse$ of the target and control groups are compared through a two-tailed $t$-test. Data outside of three standard deviations of the mean are eliminated.
The results are presented in \fref{experimntal_results}.   

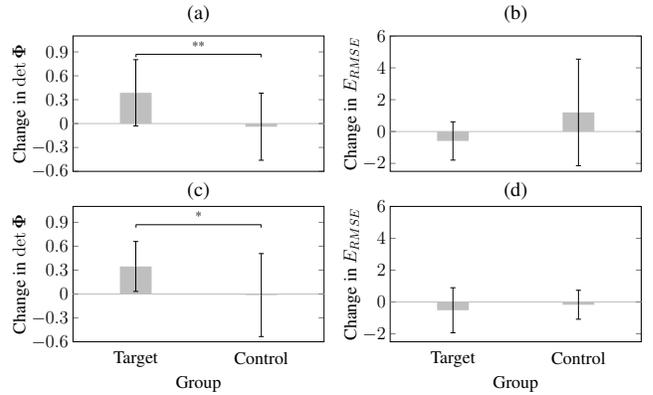
\begin{figure}[t!]
\centering
\begin{tikzpicture}[scale=0.48,trim left=-0.7cm]
      \begin{axis}[
      width=0.48\textwidth,
      height=0.3\textwidth,
      ylabel={Change in $\det\bPhi$},
      ylabel shift = -5pt,
      ymin=-0.6, 
      ymax=1.1,
      major x tick style = transparent,
      ybar=2*\pgflinewidth,
      bar width=25pt,
      xtick={1,2},
      xticklabels={},
      scaled y ticks = false,
      enlarge x limits=0.50,
      ytick={-0.6,-0.3,0.0, 0.3, 0.6, 0.9},
        extra y ticks={0},
        extra y tick labels={},
        extra tick style={
            grid=major,
        },
        label style={font=\Large},tick label style={font=\Large},
  ]
    
    \addplot[style={fill=gray!50,draw=none,},error bars/.cd, y dir=both, y explicit]
          coordinates {
          (1,0.386569619579299) +- (0,0.415541800199779)
          (2,-0.039592274043587) +- (0,0.420986802745297)};
    \draw [arrows={Bar[left]-Bar[right]}, ] 
        (0,147)  -- node[midway, above]{**} ++(100,0);
  \end{axis}
  \begin{axis}[
  width=0.48\textwidth,
      height=0.3\textwidth,
    axis x line*=top,
      xlabel={\ref{i:exp2_a}},
      ymin=-0.6, 
      ymax=1.1,
      major x tick style = transparent,
      major y tick style = transparent,
      xtick={1,2},
      xticklabels={},
      scaled y ticks = false,
      enlarge x limits=0.50,
      ytick={-0.6,-0.3,0.0, 0.3, 0.6, 0.9},
      yticklabels={},
      axis line style = { draw = none },
      label style={font=\LARGE},tick label style={font=\LARGE},
  ]
  \end{axis}
  \end{tikzpicture}
\begin{tikzpicture}[scale=0.48,trim left=-0.7cm]
      \begin{axis}[
      width=0.48\textwidth,
      height=0.3\textwidth,
      xlabel={},
      ylabel={Change in $\rmse$},
      ylabel shift = -5pt,
      ymin=-2.5, 
      ymax=6.0,
      major x tick style = transparent,
      ybar=2*\pgflinewidth,
      bar width=25pt,
      xtick={1,2},
      xticklabels={},
      scaled y ticks = false,
      enlarge x limits=0.50,
      ytick={-2.0,  0.0, 2.0, 4.0, 6.0},
        extra y ticks={0},
        extra y tick labels={},
        extra tick style={
            grid=major,
        },
        label style={font=\Large},tick label style={font=\Large},
  ]
    
    \addplot[style={fill=gray!50,draw=none,},error bars/.cd, y dir=both, y explicit]
          coordinates {
          (1,-0.591004752429438) +- (0,1.20280613812143)
          (2,1.20176995682356) +- (0,3.34435635358813)};
  \end{axis}
  \begin{axis}[
  xlabel={\ref{i:exp2_b}},
  width=0.48\textwidth,
      height=0.3\textwidth,
    axis x line*=top,
      ymin=-2.5, 
      ymax=6.0,
      major x tick style = transparent,
      major y tick style = transparent,
      xtick={1,2},
      xticklabels={},
      scaled y ticks = false,
      enlarge x limits=0.50,
      ytick={-2.0,  0.0, 2.0, 4.0, 6.0},
      yticklabels={},
      axis line style = { draw = none },
      label style={font=\LARGE},tick label style={font=\Large},
  ]
  \end{axis}
  \end{tikzpicture}

  \vspace{-1mm}

\begin{tikzpicture}[scale=0.48,trim left=-0.7cm]
      \begin{axis}[
      xlabel={Group},
      ylabel={Change in $\det\bPhi$},
      ylabel shift = -5pt,
      width=0.48\textwidth,
      height=0.3\textwidth,
      ymin=-0.6, 
      ymax=1.1,
      major x tick style = transparent,
      ybar=2*\pgflinewidth,
      bar width=25pt,
      xtick={1,2},
      xticklabels={Target, Control},
      scaled y ticks = false,
      enlarge x limits=0.50,
      ytick={-0.6,-0.3,0.0, 0.3, 0.6, 0.9},
        extra y ticks={0},
        extra y tick labels={},
        extra tick style={
            grid=major,
        },
        label style={font=\Large},tick label style={font=\Large},
  ]
    
    \addplot[style={fill=gray!50,draw=none,},error bars/.cd, y dir=both, y explicit]
          coordinates {
          (1,0.346026741798924) +- (0,0.313407061873178)
          (2,-0.013313476875) +- (0,0.521055663127769)};
    \draw [arrows={Bar[left]-Bar[right]}, ] 
        (0,147)  -- node[midway, above]{*} ++(100,0);
  \end{axis}
  \begin{axis}[
  width=0.48\textwidth,
      height=0.3\textwidth,
    axis x line*=top,
      xlabel={\ref{i:exp2_c}},
      xlabel shift = -5pt,
      ymin=-0.6, 
      ymax=1.1,
      major x tick style = transparent,
      major y tick style = transparent,
      xtick={1,2},
      xticklabels={},
      scaled y ticks = false,
      enlarge x limits=0.50,
      ytick={-0.6,-0.3,0.0, 0.3, 0.6, 0.9},
      yticklabels={},
      axis line style = { draw = none },
      label style={font=\LARGE},tick label style={font=\Large},
  ]
  \end{axis}
  \end{tikzpicture}
\begin{tikzpicture}[scale=0.48,trim left=-0.7cm]
      \begin{axis}[
      width=0.48\textwidth,
      height=0.3\textwidth,
      xlabel={Group},
      ylabel={Change in $\rmse$},
      ylabel shift = -5pt,
      ymin=-2.5, 
      ymax=6.0,
      major x tick style = transparent,
      ybar=2*\pgflinewidth,
      bar width=25pt,
      xtick={1,2},
      xticklabels={Target, Control},
      scaled y ticks = false,
      enlarge x limits=0.50,
      ytick={-2.0,  0.0, 2.0, 4.0, 6.0},
        extra y ticks={0},
        extra y tick labels={},
        extra tick style={
            grid=major,
        },
        label style={font=\Large},tick label style={font=\Large},
  ]
    
    \addplot[style={fill=gray!50,draw=none,},error bars/.cd, y dir=both, y explicit]
          coordinates {
          (1,-0.519163468676601) +- (0,1.41310114088459)
          (2,-0.174742116933333) +- (0,0.907923904782788)};
  \end{axis}
  \begin{axis}[
  xlabel={\ref{i:exp2_d}},
  xlabel shift = -5pt,
  width=0.48\textwidth,
      height=0.3\textwidth,
    axis x line*=top,
      ymin=-2.5, 
      ymax=6.0,
      major x tick style = transparent,
      major y tick style = transparent,
      xtick={1,2},
      xticklabels={},
      scaled y ticks = false,
      enlarge x limits=0.50,
      ytick={-2.0,  0.0, 2.0, 4.0, 6.0},
      yticklabels={},
      axis line style = { draw = none },
      label style={font=\LARGE},tick label style={font=\Large},
  ]
  \end{axis}
  \end{tikzpicture}
\caption{Difference in \cl{\item\label{i:exp2_a} teaching quality ($\det\bPhi$) and \item\label{i:exp2_b} learning error between \ref{i:phase1} and \ref{i:phase3}. Panels \item\label{i:exp2_c} and \item\label{i:exp2_d} show the differences in the same quantities between \ref{i:phase2} and \ref{i:phase6}, respectively.}}
\label{f:experimntal_results}
\end{figure}

First, the effectiveness of guidance using the index \eref{score} is examined. For this, the change in the quality of teaching and learner performance for \ref{i:s1} in \ref{i:phase1} and \ref{i:phase3} (\ie the non-guided and guided phases) of the target and control group are compared. Looking at \fref{experimntal_results}\ref{i:exp2_a}, it can be seen that the increase in the determinant of $\bPhi$ (\ie the quality of demonstrations) from \ref{i:phase1} to \ref{i:phase3} is significantly higher in the target group ($0.39\pm0.42$) than in the control group ($-0.04\pm0.42$, $t(30)=2.790$, $p=0.009$). This is reflected in a decrease in $\rmse$(See \fref{experimntal_results}\ref{i:exp2_b}), although this is not statistically significant ($t(29)=1.944$, $p=0.062$). The results show that guidance based on the indicators derived from \gls{mt} enables novice human teachers to produce high quality demonstrations that lead to high learner performance.

Next, the extent to which participants \emph{retain} improved teaching ability after guidance is examined. For this, the change in the quality of demonstrations and $\rmse$ of teaching \ref{i:s1} between \ref{i:phase1} and \ref{i:phase5} of the target and control groups are compared, and there was no statistically significant difference between the two groups. The target group's change in the quality of demonstrations (determinant of $\bPhi$) is $0.03\pm0.44$ and that of the $\rmse$ is $-0.01\pm1.46$. This suggests little retention of improved teaching skill. 

Finally, the extent to which improvements in participants' ability to teach one motor skill \emph{generalises} to another is evaluated. For this, the change in the quality of teaching and learner performance of teaching \ref{i:s2} in \ref{i:phase2} and \ref{i:phase6} (\ie the change in teaching ability for \ref{i:s2}, before and after guidance in teaching \ref{i:s1}) of the target and control group are compared. As can be seen in \fref{experimntal_results}\ref{i:exp2_c}, the increase in the quality of demonstrations (determinant of $\bPhi$) from \ref{i:phase2} to \ref{i:phase6} of the target group ($0.35\pm0.31$) is significantly higher than of the control group which stays at approximately the same ($-0.01\pm0.52$, $t(30) = 2.289$, $p=0.029$). This is reflected in the $\rmse$ for the two groups ($-0.52\pm1.41$ and $-0.17\pm0.91$, respectively)(See \fref{experimntal_results}\ref{i:exp2_d}). These results suggest that just one guided session can lead to an improvement in teaching ability across different skills. 

\section{CONCLUSIONS}
\label{s:conclusion}
This paper designs a machine teaching problem for teaching motor skills to dynamically controlled robots through {\Gls{lfd}}, presents a condition for informative demonstration data using {\acrlong{mt}} and validates the use of the resulting teaching quality index to guide and train human teachers. Evaluation of the latter in teaching a dynamic motor skill to a simple robot suggest that its use can enable novice teachers to improve their selection of demonstrations.
Moreover, there is evidence that guidance enhances high-level learning to select good demonstrations to teach new motor skills. 


In future work, this work may be extended to \il{\item teach more complex dynamic motor skills which require higher degrees of freedom robots or learning algorithms where no closed-form solutions are available and \item evaluate the effectiveness of different types of guidance and training methods using the \gls{mt} framework.}

\addtolength{\textheight}{-5cm}   



\cleardoublepage


{\scriptsize\printbibliography}%

\end{document}